\documentclass[runningheads]{llncs}

\usepackage{eccv}

\usepackage{eccvabbrv}

\usepackage{graphicx}
\usepackage{booktabs}

\usepackage[accsupp]{axessibility}  

\usepackage{hyperref}

\usepackage{orcidlink}

\usepackage{multirow}
\usepackage{bbding}

\begin{document}

\title{Sparse Beats Dense: Rethinking Supervision in Radar-Camera Depth Completion}

\titlerunning{Sparse Beats Dense}

\author{Huadong Li$^{*,}$\inst{1} \and
Minhao Jing$^{*,}$\inst{1} \and
Wang Jin\inst{3} \and
Shichao Dong\inst{1} \and
Jiajun Liang\inst{1} \and
Haoqiang Fan\inst{1} \and
Renhe Ji$^{\dag,}$\inst{2}
}

\authorrunning{Li.~ et al.}

\institute{MEGVII Technology \and  Fvidar Inc. 
\and
The University of Hong Kong\\
\email{ \{lihuadong, liangjiajun, fanhaoqiang\}@megvii.com, renhe.ji@fvidar.com} 
}

\def\footnotesymbollist{%
  \symbol{$*$}%
  \symbol{$\dag$}%
}

{
	\renewcommand{\thefootnote}{\fnsymbol{footnote}}
     \footnotetext[1]{Equal contribution}
	\footnotetext[4]{Corresponding author}
}

\maketitle

\begin{abstract}
It is widely believed that sparse supervision is worse than dense supervision in the field of depth completion, but the underlying reasons for this are rarely discussed.
To this end, we revisit the task of radar-camera depth completion and present a new method with \textbf{\textit{sparse LiDAR supervision}} to outperform previous \textbf{\textit{dense LiDAR supervision}} methods in both accuracy and speed.
Specifically, when trained by sparse LiDAR supervision, depth completion models usually output depth maps containing significant stripe-like artifacts.
We find that such a phenomenon is caused by the implicitly learned positional distribution pattern from sparse LiDAR supervision, termed as \textit{LiDAR Distribution Leakage} (LDL) in this paper.
Based on such understanding, we present a novel \textit{Disruption-Compensation} radar-camera depth completion framework to address this issue.
The \textit{Disruption} part aims to deliberately disrupt the learning of LiDAR distribution from sparse supervision, while the \textit{Compensation} part aims to leverage 3D spatial and 2D semantic information to compensate for the information loss of previous disruptions.
Extensive experimental results demonstrate that by reducing the impact of LDL, our framework with \textit{\textbf{sparse supervision}} outperforms the state-of-the-art \textit{\textbf{dense supervision}} methods with \textbf{11.6$\mathbf{\%}$ improvement in Mean Absolute Error (MAE)} and \textbf{$\mathbf{1.6 \times}$ speedup in Frame Per Second (FPS)}.
The code is available at \href{https://github.com/megvii-research/Sparse-Beats-Dense}{\textit{https://github.com/megvii-research/Sparse-Beats-Dense}}.
\end{abstract}

\section{Introduction}
\label{sec:intro}

Understanding the three-dimensional (3D) structure of our surrounding scenes can support a variety of spatial tasks, such as perception \cite{chen2023futr3d,huang2023tri}, planning \cite{kim2021eagermot,qureshi2019motion}, \emph{et al.}
To achieve this, agents typically need to be equipped with multiple sensors, such as cameras, and radars to obtain dense depth information.
In this paper, we focus on the radar-camera depth completion task \cite{singh2023depth,long2021radar}, which aims to obtain a dense and accurate depth map by leveraging a single image and an extremely sparse radar point cloud frame (50 to 80 points).

\begin{figure}[htbp]
    \centering
    \includegraphics[width=0.99\linewidth]{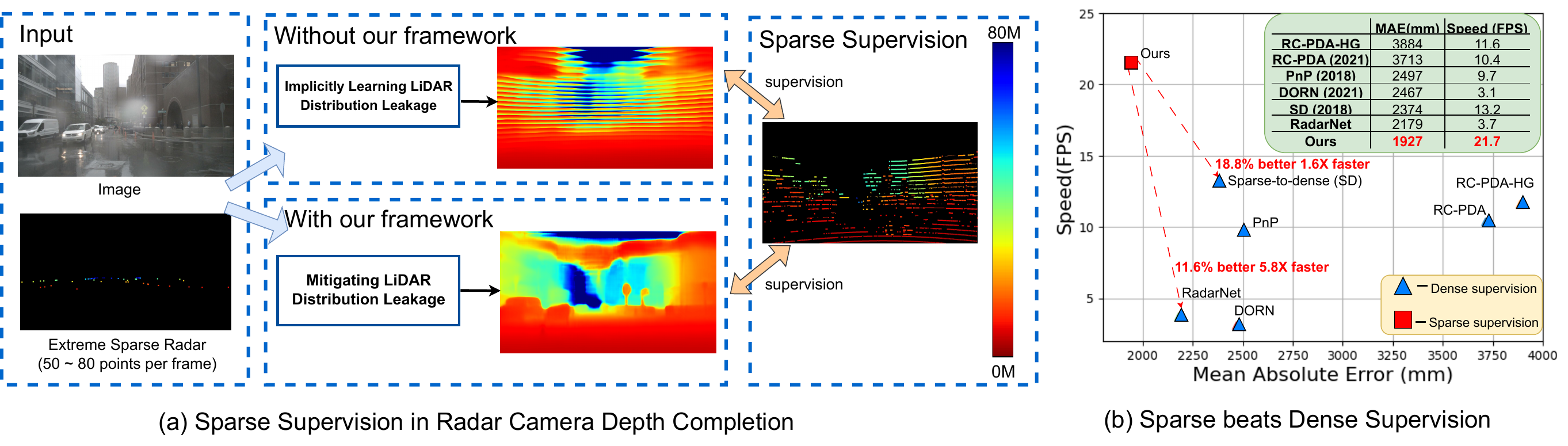}
    \caption{
    (a) Without our proposed framework, directly exploiting \textbf{\textit{sparse supervision}} of a single LiDAR frame leads to stripe-like scanning patterns in outputs.
    However, our proposed \textit{Disruption-Compensation} radar-camera depth completion framework mitigates this issue and relights the sparse supervision for this task.
    (b) Our proposed framework under \textbf{\textit{sparse supervision}} outperforms state-of-the-art \textbf{\textit{dense supervision}} methods, with \textbf{$\mathbf{11.6\%}$ improvement in MAE (Mean Absolute Error)} and \textbf{$\mathbf{1.6 \times}$ speedup in FPS (Frame Per Second)}.
    }
\label{fig:visual}
\end{figure}%

\begin{figure}[t]
    \centering
    \includegraphics[width=0.9\textwidth]{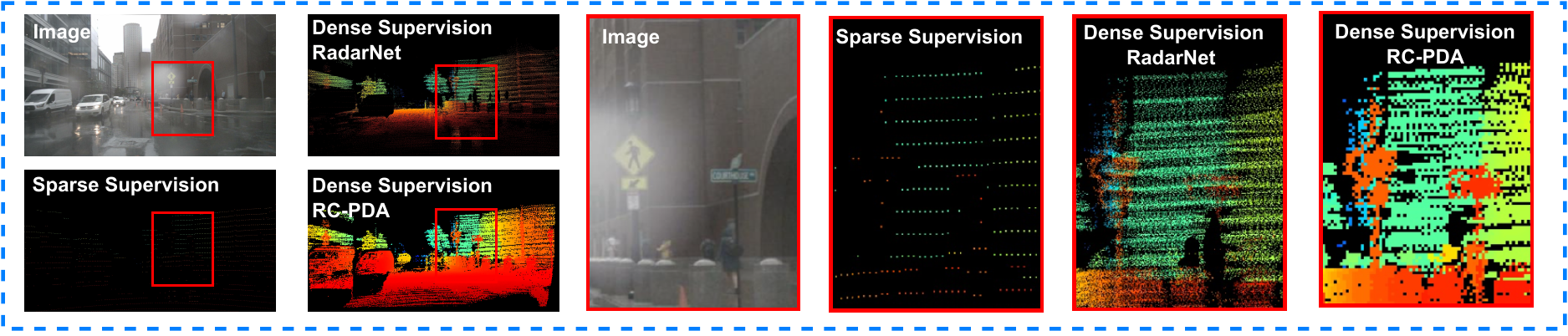}
    \caption{
    Multi-frame dense supervision is noisy.
    As shown in the red box, the depth supervision of the traffic signs and the background buildings appears messy and inaccurate due to inter-frame noises, which will confuse the depth completion models under dense supervision \cite{singh2023depth,long2021radar}.
    }
    \label{fig:noise}
\end{figure}

In this field, due to the lack of dense depth ground truth (GT), a single frame of sparse LiDAR is expected to be used as the ground truth for training and testing. 
However, recent methods \cite{long2021radar,singh2023depth} have found that training with such sparse supervision leads to severe stripe-like scanning pattern artifacts in outputs, as depicted in the upper-middle section of Fig. \ref{fig:visual} \textcolor{red}{(a)}.
These stripe-like scanning pattern artifacts signify that the depth predictions beyond the supervised points are significantly inaccurate, rendering them unsuitable for downstream applications. 
As a workaround, current methods \cite{singh2023depth,wang2018plug,ma2018sparse,gasperini2021r4dyn,lo2021depth,long2021radar} have shifted their focus towards generating dense supervision data from sparse supervision data, employing techniques such as multi-frame stacking \cite{long2021radar,gasperini2021r4dyn,singh2023depth} and interpolation \cite{lo2021depth,wang2018plug,ma2018sparse}. 
However, these approaches unavoidably introduced noises to the ground truth depth maps, such as inter-frame noises and interpolation errors.
As shown in Fig. \ref{fig:noise}, the multi-frame stacked dense supervision employed in RadarNet \cite{singh2023depth} and RC-PDA \cite{long2021radar} results in the depth map intersections of traffic signs and background buildings due to inter-frame noises. 
This inaccurate dense supervision can inherently degrade the accuracy of current models.

Therefore, we turn our concentration back to sparse supervision in the radar-camera depth completion task, so as to essentially abandon the 
\textit{cumbersome} procedures of generating \textit{noised} dense supervision data for training preparations.
Specifically, under sparse LiDAR supervision, we identify that models producing stripe-like artifacts in output depth maps are attributed to what we term as the \textit{LiDAR Distribution Leakage} (LDL).
As illustrated in Fig. \ref{fig:visual} \textcolor{red}{(a)}, we indicate LiDAR distribution as the stripe-like positional distribution pattern in the ground truth 2D projected LiDAR depth maps.
This unique distribution of sparse LiDAR data tends to make models put less effort into learning the depth representations beyond sparse supervised points, resulting in capturing incomplete 3D information across the entire space.
Please see Sec. \ref{DCV} for details.

\begin{figure*}[t]
    \centering
    \includegraphics[width=0.99\textwidth]{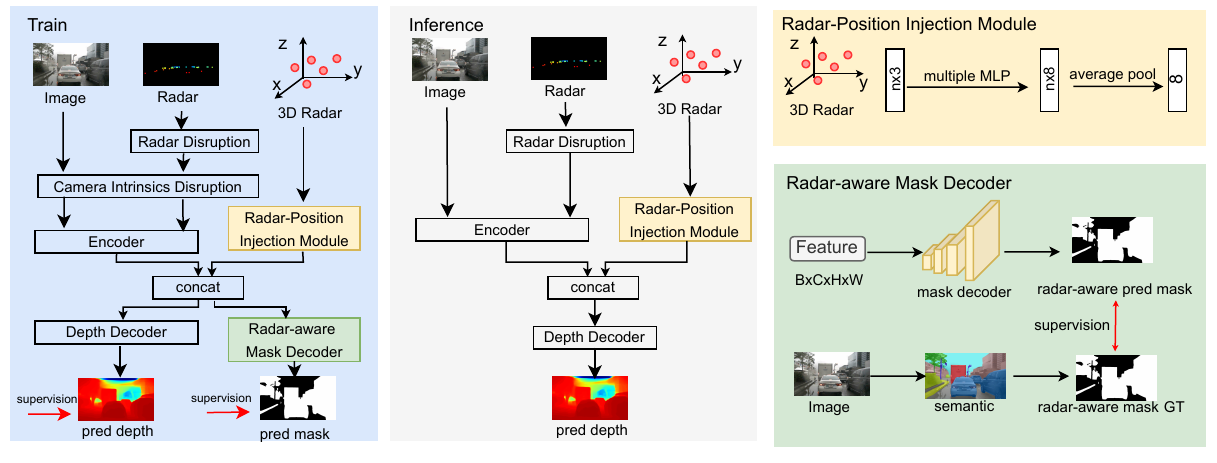}
    \caption{The architecture of \textit{Disruption-Compensation} radar-camera depth completion framework. 
    In the \textit{Disruption} part, we propose Camera Intrinsics Disruption and Radar Disruption to mitigate the impact of LiDAR Distribution Leakage.
    In the \textit{Compensation} part, we design the Radar-Position Injection Module and the Radar-aware Mask Decoder to compensate for the information loss in the previous disruption process.
    Notably, the Radar-aware Mask Decoder is only used during training, bringing no additional computational cost in the inference phase.
    }
    \label{fig:Method}
\end{figure*}

Based on this key insight, as shown in Fig. \ref{fig:Method}, we design a novel \textit{Disruption-Compensation} radar-camera depth completion framework to handle the issue of LDL. 
The \textit{Disruption} part aims to mitigate the impact of LDL by disrupting the learning of the positional distribution pattern from LiDAR data.
It consists of two operations: Camera Intrinsics Disruption and Radar Disruption, which involve applying spatial augmentations on input data (image and radar) and supervision data (LiDAR), as well as height densifications on radar data.
The \textit{Compensation} part seeks to offset the information loss incurred by the \textit{Disruption} part and consists of two modules: the Radar-aware Mask Decoder and the Radar-Position Injection Module. 
The Radar-aware Mask Decoder aims to generate a binary mask that indicates the locations of 12 classes in the image, such as cars, people, traffic lights, etc.
Since most response points of 2D radar images are within these mask regions due to the radar's physical characteristics, we utilize this mask as auxiliary supervision during training to compensate for disruptions in radar position information.
The Radar-Position Injection Module is a small Multi-Layer Perceptron (MLP), which directly extracts information from 3D radar points.
This module guides the model to capture rich information in the 3D space, thereby enhancing its performance in the depth completion task.

Extensive experimental results show that our model with \textit{\textbf{sparse supervision}} accurately predicts the depth of objects in images, successfully outperforming state-of-the-art (SOTA) \textit{\textbf{dense supervision}} methods with \textbf{11.6$\mathbf{\%}$ improvement in MAE (Mean Absolute Error)}, and \textbf{run $\mathbf{1.6 \times}$ faster in FPS (Frame Per Second)}, as shown in Fig. \ref{fig:visual} \textcolor{red}{(b)}. 
The contributions  of the paper are summarized as follows:

\noindent $\bullet$ We revisit the sparse LiDAR supervision in the radar-camera depth completion task and introduce the \textit{LiDAR Distribution Leakage} (LDL), which sheds light on why sparse LiDAR supervision leads to unusable depth maps. 

\noindent $\bullet$ We propose a novel \textit{Disruption-Compensation} radar-camera depth completion framework to tackle the issue of LDL.
This sparse-supervised framework outperforms other SOTA dense supervision methods in both accuracy and speed.

\noindent $\bullet$ We conduct extensive experiments to verify the impact of LDL and demonstrate the effectiveness of our proposed framework.

\section{Related Work}
\label{sec:Related}

\noindent \textbf{Radar-Camera Depth Completion.}
Radar-camera depth completion uses extremely sparse radar point clouds and camera images to obtain dense depth maps with sparse LiDAR point clouds supervision.
Existing methods \cite{long2021radar,singh2023depth} have found that training based on single-frame LiDAR (sparse supervision) leads to severe stripe-like scanning pattern artifacts.
As a workaround, previous methods \cite{long2021radar,singh2023depth, lo2021depth} typically employed multi-frame accumulation or interpolation to obtain dense supervision.
While these hand-crafted dense LiDAR supervision methods alleviated the problem of  stripe-like artifacts, they inevitably introduced noises such as inter-frame noises and interpolation errors.
In contrast, our proposed \textit{Disruption-Compensation} framework mitigates the scanning pattern problem under sparse supervision and achieves high performance and efficiency. 

\noindent \textbf{LiDAR-Camera Depth Completion.}
This task \cite{lin2022dynamic,rho2022guideformer,yan2022rignet,qiu2019deeplidar,cheng2018depth,hu2021penet,imran2021depth} aims to obtain dense depth maps given LiDAR data and camera images, which shares similar targets to our radar-camera depth completion task. 
However, compared to LiDAR input data, radar input point clouds are noisy and extremely sparse with low resolutions, rendering these methods unsuitable for solving our task.

\noindent \textbf{Implicit Neural Representation.}
As a direct yet efficient method, studies in this field usually proposed to map the coordinates to visual representations with a Multi-Layer Perceptron (MLP), which was commonly utilized to encode 3D objects \cite{park2019deepsdf,chen2019learning,mescheder2019occupancy}, 2D images \cite{liu2022petr,philion2020lift,hu2019meta,sitzmann2020implicit}, and 3D scenes \cite{mildenhall2021nerf,chabra2020deep}.
Inspired by the above methods, we build the Radar-Position Injection Module in our framework to directly extract information from 3D radar points and guide the model to obtain positional knowledge in the 3D space.

\begin{figure}[t]
    \centering
    \includegraphics[width=0.99\textwidth]{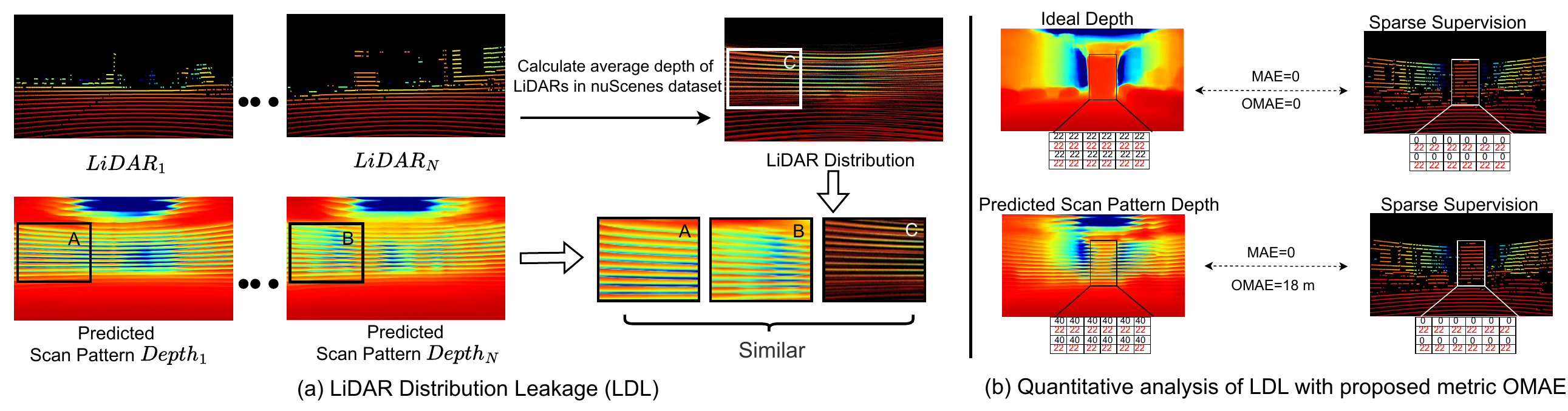}
    \caption{(a) \textbf{LiDAR Distribution Leakage (LDL).} We denote the LiDAR distribution as the stripe-like positional distribution pattern in the ground truth 2D projected LiDAR depth maps. We notice that although certain regions (\emph{e.g.}, region A and B) on LiDAR images contain no depth values for supervision, the depth completion model still fulfilled depth values for these regions, following the stripe-like artifacts pattern of the LiDAR distribution (\emph{e.g.}, region C). Such results qualitatively indicate the leakage of LiDAR distribution during training. (b) Quantifying the impact of LDL with our newly proposed metric OMAE. Between the truck region in the ideal output dense map and the scan-pattern output depth map, the MAE metric used in \cite{singh2023depth} returns the same value 0 since it only considers limited supervised positions of sparse LiDAR data (\emph{e.g.}, depth values marked in red), failing to reflect the impact of LDL. In comparison, OMAE considers object-level depth differences, which takes the stripe-like areas of each object into account, enabling the reflection about the impact of LDL.}
    \label{fig:mono}
\end{figure}

\section{LiDAR Distribution Leakage}
\label{DCV}
As shown in previous studies \cite{long2021radar,singh2023depth}, directly utilizing sparse LiDAR data as supervision leads to stripe-like artifacts in the output depth maps of depth completion models.
In this section, we dive deeply into the causes of such a phenomenon and introduce the concept of \textit{LiDAR Distribution Leakage} (LDL), which sheds new light on relighting sparse supervision for this task.

Specifically, we indicate the LiDAR distribution as the stripe-like positional distribution of the data collected by the LiDAR sensor when projected to the 2D reference image.
This distribution pattern roughly depends on the characteristics of the LiDAR sensor (\emph{e.g.}, the scanning pattern) and the projection transformation of LiDAR data onto the 2D reference image.
This projection transformation process further depends on the intrinsic camera parameters and the relative extrinsic parameters between the camera and the LiDAR sensor.
Since the recent training set \cite{caesar2020nuscenes} is collected using the same paired LiDAR-camera device, all data inevitably features the same LiDAR distribution, which makes it easily captured by depth completion models during the training phase.
Such implicitly-learned yet undesired knowledge then results in the predictions of depth maps containing stripe-like artifacts.
We term this phenomenon as the \textit{LiDAR Distribution Leakage} (LDL) in this paper.
For a better visual understanding, as shown in Fig. \ref{fig:mono} \textcolor{red}{(a)}, we calculated the average depth value for every location in the 2D projected LiDAR images in the nuScenes \cite{caesar2020nuscenes} training dataset, which shows remarkable resemblances to the output depth maps in terms of stripe-like artifacts.
In particular, even though certain regions (\emph{e.g.}, region A and B) on the 2D projected LiDAR images contain no depth values for sparse supervision, the depth completion model still filled up depth values on the predicted depth maps, following the stripe-like artifacts pattern of the LiDAR distribution (\emph{e.g.}, region C).
Such results provide a qualitative understanding that the depth completion model under sparse LiDAR supervision implicitly learned the LiDAR distribution from training data. 
To further comprehensively analyze this phenomenon, we then design a new metric to quantify the impact of LDL.

\subsection{Quantifying the LiDAR Distribution Leakage}
Previous metrics for evaluating the performance of depth completion models, such as MAE and RMSE used in \cite{singh2023depth}, are insufficient to quantify the influence of LiDAR Distribution Leakage.
Specifically, these metrics only calculate the depth differences between the sparse ground truth LiDAR points and the corresponding positions in the predicted depth maps.
They overlook the predicted depth values in positions without ground truth, leading to biased evaluations, especially for quantifying the significance of stripe-like patterns in depth maps.

Therefore, to properly quantify the impact of LDL, we extend the previous MAE metric and propose the Object-level MAE (OMAE) metric.
Intuitively, the stripe-like pattern in the predicted depth map violates human's understanding of semantic concepts, \emph{i.e.}, the depth region of an object should be smooth instead of encompassing depth streaks.
To this end, our proposed OMAE metric aims to measure the object-level depth differences between the predicted depth map and the ground truth 2D projected LiDAR depth map.
Specifically, for the ground truth data, we propose to calculate the average depth value \textit{within the sparse points of each object} in the 2D projected LiDAR image.
As for the predicted depth map, we propose to calculate the average depth value \textit{within each object region}.
When the predicted depth map is the ideal dense depth map with no stripe-like artifacts, the differences between the predicted depth map and the LiDAR ground truth in terms of this object-level depth should be minimal.
In contrast, if the strip-like pattern appears on the predicted depth map, such object-level depth differences shall be large, owing to the completely wrong depth predictions in certain parts of object regions.
For a better visual understanding, we provide a comparison to the MAE metric in Fig. \ref{fig:mono} \textcolor{red}{(b)}.
Mathematically, OMAE is calculated as follows,
\begin{equation}
    OMAE = \sum_{o \in \mathcal{O}}\frac{|\Omega^o_{gt}|}{|\Omega_{gt}|}  |\frac{1}{|\Omega^o_{pred}|} \sum_{x \in \Omega^{o}_{pred}} \hat{d}(x)- \frac{1}{|\Omega^o_{gt}|} \sum_{x \in \Omega^{o}_{gt}} d_{gt}(x)|
\end{equation}
where $\Omega^o_{pred}$ denotes the object region in the predicted depth map for object $o$ of the whole object set $\mathcal{O}$; $\Omega^o_{gt}$ denotes the sparse supervised points in the ground truth 2D projected LiDAR image for object $o$ of the whole object set $\mathcal{O}$; $\Omega_{gt}$ denotes all the supervised points in the ground truth 2D projected LiDAR depth map; $\hat{d}(\cdot)/d_{gt}(\cdot)$ denotes the depth value in the predicted depth map/ground truth 2D projected LiDAR image.
In implementation, we obtained each object region by exploiting ViT-Adapter \cite{chen2022vision} for the nuScenes \cite{caesar2020nuscenes} dataset\footnote{These object-level mask annotations will be released as well, along with the code.}.

With this metric, we are now capable of quantitatively measuring the significance of LDL, where a severer stripe-like pattern in the predicted depth map results in a larger value of OMAE and a more accurate depth map with less severe stripe-like artifacts leads to a smaller value of OMAE.
In this way, we can further quantitatively measure the effectiveness of our proposed framework in mitigating the impact of LDL.

\begin{figure}[t]
    \centering
    \includegraphics[width=0.99\textwidth]{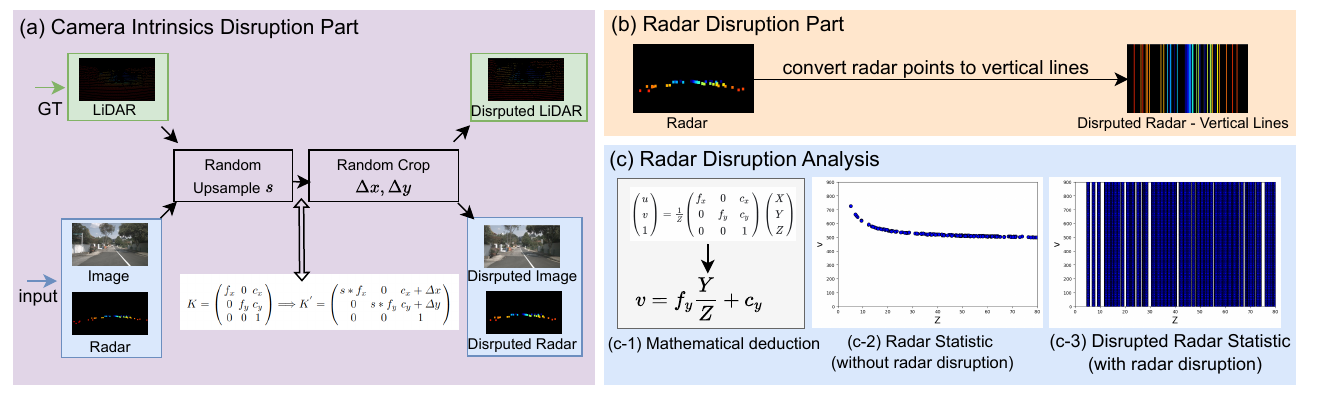}
    \caption{The \textit{Disruption} part: Camera Intrinsics Disruption and Radar Disruption. (a) In the Camera Intrinsics Disruption process, for the input image/radar data and the ground truth LiDAR data, we propose to randomly upsample them with a ratio $s$ and randomly crop them with the offset $\Delta x$ and $\Delta y$. This process is equivalent to disrupting camera intrinsics from $K$ to $K^{\prime}$. Here $f_x$, $f_y$, $c_x$, and $c_y$ are camera intrinsics. (b) In the Radar Disruption process, we propose to extend the radar points on the input 2D radar image into vertical lines. (c) We provide analysis results here to justify the reasons for the Radar Disruption process. Here Fig. (c-1)/(c-2) both denote that camera intrinsics $f_y$ and $c_y$ can be roughly inferred based on a simple inverse proportional function, which can be easily captured by models. Fig. (c-3) shows that after applying the Radar Disruption, $f_y$ and $c_y$ can no longer be easily obtained through a simple curve function, hindering models from capturing camera intrinsics. See Sec. \ref{sec:Radar_Disruption} for details.}
     \label{fig:distruption}
\end{figure}

\section{Disruption - Reducing LiDAR Distribution Leakage}
In this section, as shown in Fig. \ref{fig:distruption}, we first present disruption methods to mitigate the impact of LDL, \emph{i.e.}, Camera Intrinsics Disruption and Radar Disruption, expecting to improve the model performance under sparse LiDAR supervision.
\subsection{Camera Intrinsics Disruption}
Based on the understanding of the LiDAR Distribution Leakage (LDL), one can mitigate its impact by disrupting the stripe-like positional distribution of LiDAR data when projected to the 2D reference image.
Considering the three main factors that impact this distribution pattern \emph{i.e.}, the characteristics of the LiDAR sensor, the relative extrinsic parameters between the camera and the LiDAR sensor, and the intrinsic camera parameters, we propose the Camera Intrinsics Disruption to break this LiDAR distribution pattern for the ease of implementations.
In this way, each 2D projected LiDAR image is expected to be obtained with different camera intrinsics, presenting obstacles to depth completion models to implicitly learn the LiDAR distribution of the dataset.

Specifically, we propose to disrupt the camera intrinsics with spatial augmentations on input data (image and radar) and supervision data (LiDAR).
As shown in the left part of Fig. \ref{fig:distruption}, we propose to randomly upsample and randomly crop the data, which are equivalent to disrupting camera intrinsics \cite{hartley2003multiple}.
To verify the effectiveness of these operations, we conduct experiments in the nuScenes dataset \cite{caesar2020nuscenes}.
As shown in Tab. \textcolor{red}{(b-1)} within Fig. \ref{fig:generalization}, we show that exploiting these operations results in performance gain in OMAE, which demonstrates the effectiveness of the proposed disruption methods on mitigating LDL.

\subsection{Radar Disruption}
\label{sec:Radar_Disruption}
\begin{figure}[t]
    \centering
    \includegraphics[width=0.99\textwidth]{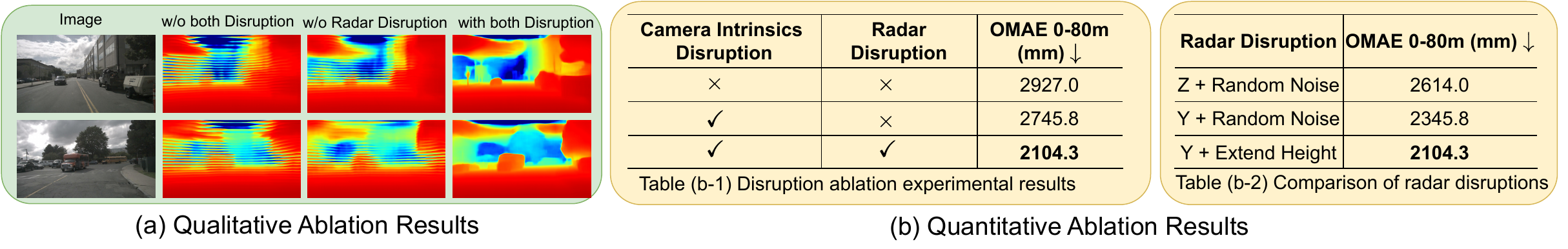}
    \caption{Ablation results on the \textit{Disruption} part our of framework. We provide both qualitative and quantitative results to demonstrate the effectiveness of the proposed two disruptions: Camera Intrinsics Disruption and Radar Disruption. Here in Tab. \textcolor{red}{(b-2)}, $Z$ represents the coordinate in the forward direction of the vehicle and $Y$ represents the coordinate in the direction perpendicular to the ground.
    }
    \label{fig:generalization}
\end{figure}

In Camera Intrinsics Disruption, we expect to make camera intrinsics vary for each input sample, so as to hinder models from learning the same camera intrinsics of the entire dataset.
Nevertheless, in this section, we show that models can still easily learn each newly disrupted camera intrinsics \textit{based on each new sample alone}, which still provides an easy path for depth completion models to generate depth maps containing stripe-like artifacts.
As shown in Fig. \ref{fig:generalization} \textcolor{red}{(a)} and Tab. \textcolor{red}{(b-1)} within Fig. \ref{fig:generalization} \textcolor{red}{(b)}, we find that previous augmentations brought 
moderate improvements and failed to completely remove the stripe-like artifacts in the predicted depth maps, which motivates us to further explore the internal causes.
To this end, we find that \textit{when given the 2D projected radar image} as input, depth completion models can still find shortcuts to implicitly learn the newly disrupted camera intrinsics \textit{per sample}.
We provide analyses as follows.


As shown in Fig. \ref{fig:distruption} \textcolor{red}{(c-1)}, given a position coordinate $(X, Y, Z)$ of a projected radar point in the camera coordinate system, where $Z$ represents the coordinate in the forward direction of the vehicle and $Y$ represents the coordinate in the direction perpendicular to the ground, then its coordinate $(u, v)$ in the screen coordinate system on the 2D projected radar image can be calculated as follows,
\begin{equation}
    u = f_x  \frac{X}{Z} + c_x,\quad v = f_y \frac{Y}{Z} + c_y
    \label{eq:uv}
\end{equation}
where $f_x$, $f_y$, $c_x$, and $c_y$ are camera intrinsics.
We then show that the values of the above camera intrinsics can be easily obtained for each input 2D projected radar image \textit{individually}, especially for $f_y$ and $c_y$. 
Specifically, considering the mechanism of a radar sensor, we can roughly consider $Y$ as a constant value for radar points in each input 2D projected radar image \cite{caesar2020nuscenes,long2021radar,singh2023depth}.
In this way, $f_y$ and $c_y$ can be easily inferred from the input 2D projected radar image if at least two points exist.
This is because $v$ and $Z$ are known per radar point, making the equation $v = f_y \frac{Y}{Z} + c_y$ contain only two variables.
In fact, $v$ and $Z$ should be roughly inversely related, which is experimentally illustrated in Fig. \ref{fig:distruption} \textcolor{red}{(c-2)}.
As a result, such a simple curve fit of $v$ and $Z$ provides an easy shortcut for depth completion models to learn the camera intrinsics of $f_y$ and $c_y$, which further helps models to capture the newly disrupted LiDAR Distribution per sample.
This newly disrupted LiDAR Distribution, though being randomly upsampled and randomly cropped in previous augmentations, can still lead to the output depth maps containing stripe-like artifacts.

To address the above issue, we propose the Radar Disruption, which aims to disrupt the input radar data, so as to mitigate the leakage of camera intrinsics.
Specifically, considering the above analyses, we propose to extend the radar points on the input 2D radar images into vertical lines, as shown in Fig. \ref{fig:distruption} \textcolor{red}{(b)}.
In this way, $v$ and $Z$ can no longer be easily fit with an inverse proportional function curve, as illustrated in Fig. \ref{fig:distruption} \textcolor{red}{(c-3)}, which causes trouble for models to learn camera intrinsics.
In experiments, as shown in Tab. \textcolor{red}{(b-1)} within Fig. \ref{fig:generalization} \textcolor{red}{(b)}, we show that Radar Disruption helped the model to achieve significant improvements in OMAE.
Besides, to gain the full picture, we also explore the effectiveness of adding noises to $Y/Z$, which is also helpful to bury camera intrinsics $f_y$ and $c_y$ in data. 
As shown in Tab. \textcolor{red}{(b-2)} within Fig. \ref{fig:generalization} \textcolor{red}{(b)}, we experimentally show that Radar Disruption, which extended the radar points on the input 2D radar images into vertical lines, obtained the most improvements in OMAE, demonstrating its superiority.
Finally, when we combined Camera Intrinsics Disruption and Radar Disruption to training data, we successfully removed stripe-like artifacts in the output depth maps, as illustrated in Fig. \ref{fig:generalization} \textcolor{red}{(a)}. 

\section{Compensation - Offsetting Positional Information Loss}

In this section, we further present the \textit{Compensation} part in our framework, which aims to offset the information loss of previous disruption operations. 
It consists of two modules: the Radar-aware Mask Decoder and the Radar-Position Injection Module. 

\subsection{Radar-aware Mask Decoder}
The Radar-aware Mask Decoder is designed to implicitly inject the radar's potential positional information. 
Its basic idea is that certain objects are more likely to be hit by radar points. Thus,
the semantic segmentation regions of these objects, containing potential position information for radar points, can serve as auxiliary supervision during training.

Specifically, radar sends electromagnetic waves \cite{skolnik1980introduction} through a transmitter. The wave hits objects in the scene, reflects back and is collected by the receiver.
Materials that reflect radar signal waves well are those with properties that have minimal absorption of these waves \cite{weinstein1988electromagnetic}, such as metals, aluminum, certain plastics \emph{et al.}
Considering the driving scene, the objects hit by the radar points are more likely to be cars, obstacles, people, traffic lights, walls, etc.
This means that, in 2D projected radar images, the majority of radar points are located within the regions of these objects. 
To this end, we select 12 categories of objects on images, including the wall, building, tree, person, car, fence, bus, truck, pole, bicycle, van, and minibike. 
Then we use the union of their semantic segmentation regions to generate the binary radar-aware masks as supervision for our Radar-aware Mask Decoder, whose goal is to generate such radar-aware masks.
The Radar-aware Mask Decoder is a type of U-Net decoder in the backbone of ResNeXt \cite{xie2017aggregated}, with a single-channel output layer following a sigmoid activation to generate the binary mask.
In experiments, we used the nuScenes \cite{caesar2020nuscenes} dataset and exploited ViT-Adapter \cite{chen2022vision} pretrained on the ADE20K \cite{zhou2019semantic} dataset to obtain these radar-aware masks as supervision.
With this auxiliary supervision for our Radar-aware Mask Decoder, we can narrow the search space of the points in the 2D projected radar images under semantic guidance, implicitly helping the model to compensate radar’s positional information.
Moreover, as illustrated in Fig. \ref{fig:Method}, the Radar-aware Mask Decoder is designed only for training, which avoids extra computational costs during the inference phase.

\subsection{Radar-Position Injection Module}
The Radar-Position Injection Module is a Multi-Layer Perceptron (MLP), which aims to directly extract positional information of 3D radar points. 
Given $n$ radar points $\mathcal{P} \in R^{n\times 3}$ in the form of
3D radar coordinates, we obtain the positional information of radar points as follows: $ F_{pj} = \Psi (\mathcal{P})$, where $F_{pj} \in  \mathbb{R}^{n \times C}$ is the injected positional feature; $\Psi$ denotes the Radar-Position Injection Module, which is a MLP.
The number of channels in each layer is (32, 64, 96, 64, 32, 8).

\subsection{Depth Completion Network}
Besides the \textit{Compensation} part, our framework also encompasses a Depth Completion Network to output depth maps, which is a multi-scale U-Net with the ResNeXt \cite{xie2017aggregated} backbone, as shown in Fig. \ref{fig:Method} (\emph{i.e.}, the Encoder and the Depth Decoder).
Its input is the multi-scale data pyramid, encompassing original images and radar data, as well as $1/2$ and $1/4$ respectively downsampled images and radar data.
The Encoder of U-Net contains 5 layers, and the number of channels in each layer is (512, 256, 128, 64, 16). 
At the end of the Encoder, there is a feature pyramid network (FPN) \cite{lin2017feature} for multi-scale feature fusion, after which a lightweight Depth Decoder is used to output depth maps, with the number of channels in each layer being (64, 16, 8, 8, 1).

\subsection{Loss Function}

The overall loss function is a weighted loss:
\begin{equation}
\label{loss}
    L = \lambda_{1} L_{G_{1}} + \lambda_{2} L_{G_{2}} + \lambda_{3} L_{G_{3}} + \lambda_{4} L_{Mask}
\end{equation}
where $L_{G_{1-3}}$ are multi-scale sparse depth supervision losses and $L_{Mask}$ is the auxiliary mask supervision loss.
$\lambda_{1-4}$ are positive scalars, where $\lambda_{1-3}$ control the weights among sparse supervision losses at different levels of the pyramid and $\lambda_{4}$ controls the weight of the auxiliary mask supervision.
$G_{1-3}$ denote different input scales, which are $(1, 1/2, 1/4)$ respectively.
The used loss function $L_{G_{1-3}}$ and $L_{Mask}$ are all Smooth L1 loss \cite{girshick2015fast}, 
\begin{equation}
\label{smoothL1}
  \textrm{smooth}_{L_1}(\rho) =
  \begin{cases}
    0.5\rho^2& \text{if } |\rho| < 1\\
    |\rho| - 0.5& \text{otherwise},
  \end{cases}
\end{equation}
where $\rho = \hat{d}(x) - d_{gt}(x)$ for $\forall x \in \Omega_{gt}$ for $L_{G_{1-3}}$, only performing supervision at each supervised position of 2D projected LiDAR images.
$\rho = \hat{seg}(x) - seg_{gt}(x)$ for $\forall x \in \Omega_{seg\_gt}$ for mask supervision $L_{Mask}$, where $\hat{seg}(\cdot)/seg_{gt}(\cdot)$  denotes the predicted segmentation mask/the ground truth segmentation mask; $\Omega_{seg\_gt}$ denotes the supervised mask regions, containing the union of 12 categories.

\section{Experiment}
\label{sec:Exp}

\begin{table}[tp]
 \caption{\textbf{Comparisons with SOTA methods in the task of radar-camera depth completion} with the nuScenes dataset, measured in OMAE, MAE and RMSE (lower is better). 
    The number in the \textit{GT} (Ground Truth) column indicates how many frames of LiDAR are used for supervision densification, while \textit{interpolation} indicates the use of interpolation algorithms for supervision densification.
    The number in the \textit{Radar Frames} column indicates how many frames of Radar data are used for input densification.
    Our method outperformed SOTA methods in both accuracy and speed.
    }
    \centering
    \renewcommand{\arraystretch}{1.2}
    \resizebox{0.99\textwidth}{!}{ 
    \begin{tabular}{l| ccc | c |ccc| ccc|c }
    \toprule
         \multirow{2}{*}{Method} & \multirow{2}{*}{GT}  & \multirow{2}{*}{Radar Frames}  & \multirow{2}{*}{Images} &  OMAE (mm) $\downarrow$ & \multicolumn{3}{|c|}{MAE (mm) $\downarrow$ }  & \multicolumn{3}{c|}{RMSE (mm) $\downarrow$} & \multirow{2}{*}{FPS} \\
         \cline{5-11} 
         & & & & \{0-80m\} & \multicolumn{3}{|c|}{\{0-50,70,80m\}} & \multicolumn{3}{c|}{\{0-50,70,80m\}} & \\
         
         \midrule
          RC-PDA \cite{long2021radar}         & 25  & 5 & 3 & 3635.4 &  2225.0 & 3326.1 & 3713.6 & 4156.5 & 6700.6 & 7692.8 & 10.4\\
          RC-PDA with HG \cite{long2021radar} & 25  & 5 & 3 & 3584.2 & 2315.7 & 3485.6 & 3884.3 & 4321.6 & 7002.9 & 8008.6 & 11.6\\
          DORN \cite{lo2021depth}             & interpolation & 15  & 1   & 2152.4 & 1926.6 & 2380.6 & 2467.7 & 4124.8 & 5252.7 & 5554.3 &  3.1\\
          R4Dyn \cite{gasperini2021r4dyn}     & 7              & 4 & 1 &  - & -  & - &  -     & - &  - & 6434.0 & - \\
          Sparse-to-dense \cite{ma2018sparse} & interpolation  & 3 & 1 & - & -  & - & 2374.0 & - &  - & 5628.0 & 13.2\\
          PnP \cite{wang2018plug}             & interpolation  & 3 & 1 &  - & -  & - & 2496.0 & - &  - & 5578.0 & 9.7\\
          RadarNet \cite{singh2023depth}      & 161 & 1  & 1 & 2165.9 & 1727.7 & 2073.2 & 2179.3 & 3746.8 & 4590.7& 4898.7 &  3.7\\
          \hline
          Ours                                & 1   & 1 & 1 & \textbf{1817.0} & \textbf{1524.5} & \textbf{1822.9} & \textbf{1927.0} & \textbf{3567.3} & \textbf{4303.6} & \textbf{4609.6} &  \textbf{21.7}  \\
           \bottomrule
    \end{tabular}}
    \label{tab:nuscnes}
\end{table}

\begin{figure*}[t]
    \centering
    \includegraphics[width=0.9\textwidth]{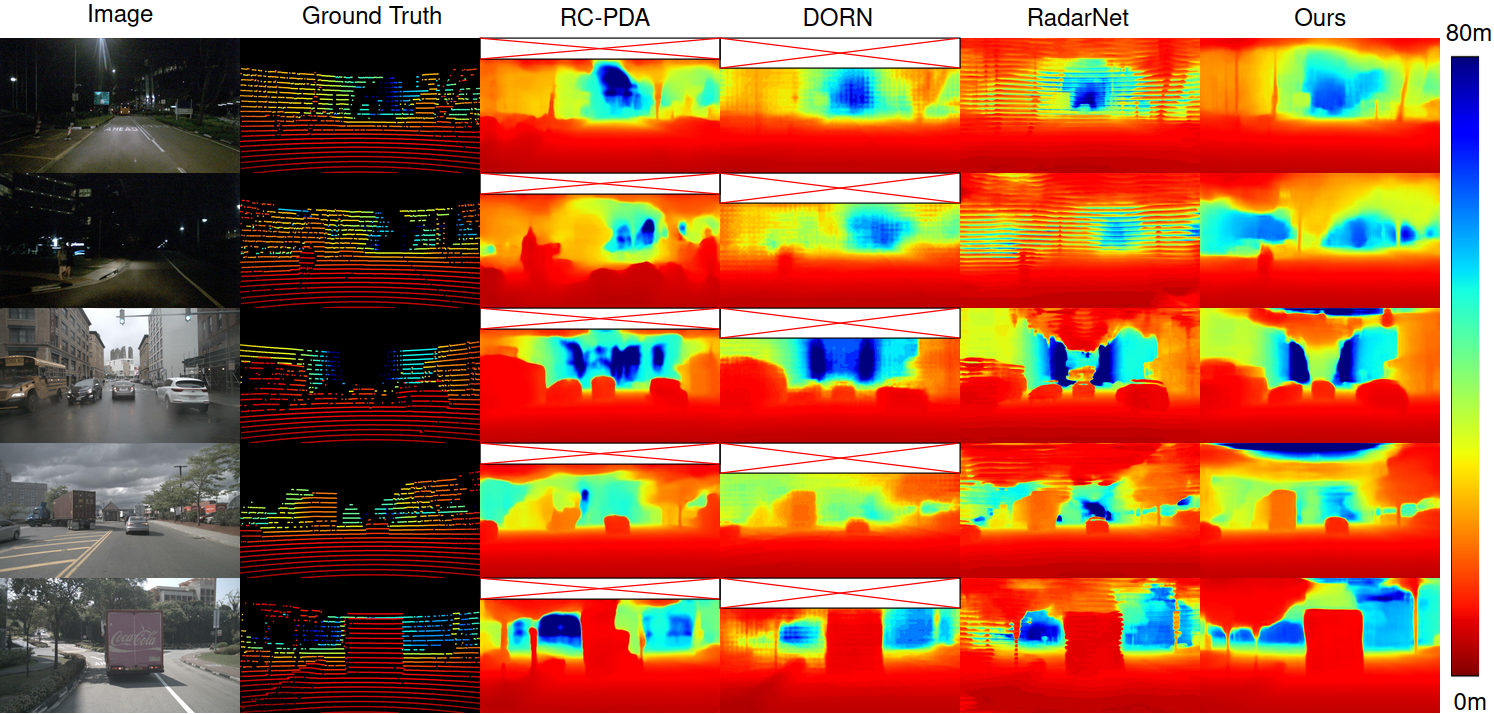}
    \caption{\textbf{Qualitative results on nuScenes test set (LiDAR best viewed in color at 10x)}. 
    Since RC-PDA \cite{long2021radar} and DORN \cite{lo2021depth} cropped the top sky area for training and inference, we filled the top area with a blank box for comparisons.
    The visualization results show that our proposed \textit{Disruption-Compensation} framework predicted artifact-free depth maps, in comparison to the scan pattern in RadarNet \cite{singh2023depth} or the grid pattern in DORN \cite{lo2021depth}.
    Besides, as shown by the regions of trucks in rows 4-5, our predicted depth maps were more accurate and smoother for the perceptions of object shapes.
    }
    \label{fig:compare}
\end{figure*}

\textbf{Dataset and Metrics.} In this paper, we performed all experiments on the nuScenes \cite{caesar2020nuscenes} dataset.
Specifically, the nuScenes dataset is a large-scale multi-modal dataset, which is composed of data
from 6 cameras, 1 LiDAR sensor and 5 radar sensors. The dataset has
1000 scenes in total and is divided into 700/150/150 scenes
as train/validation/test sets, respectively. 
Referring to RadarNet \cite{singh2023depth}, we utilized pairs of front camera and front radar from the nuScenes dataset. Our experiments were conducted using single-frame 2D radar images, without incorporating past or future continuous frames. Additionally, the ground truth for training/testing was single-frame 2D LiDAR images. 
The depth range for training and testing was 0-80 meters, with the resolution of input and output images being 900 $\times$ 1600.
As for metrics, we followed RadarNet \cite{singh2023depth}, reporting the Mean Absolute Error (MAE), and Root-Mean-Square Error (RMSE) between 0-50, 0-70 and 0-80 meters.
Besides, we also reported our newly proposed OMAE metric for further comparison.

\noindent \textbf{Implementation Details.} We trained the model using 8 RTX 2080Ti GPUs and tested the speed using 1 A10 GPU on $900 \times 1600$ image size. 
The Adam \cite{adam} optimizer was employed with an initial learning rate of 7 $\times 10^{-3}$  and a batch size of 32. 
We applied a cosine decay strategy for the learning rate scheduling. The total number of training epochs was set to 400. 
Data augmentations were used during training, including the random left-right flipping with a probability of 0.5, and brightness, saturation, and contrast adjustment.
All pyramid output depth maps were supervised, and the corresponding weight hyperparameters were $\lambda_1 = 1$, $\lambda_2 = 0.5$, $\lambda_3 = 0.25$, and $\lambda_4 = 1$.

\noindent \textbf{State-of-the-Art Comparisons.} We compared our method against different
methods \cite{singh2023depth,long2021radar,lo2021depth,gasperini2021r4dyn,lin2020depth,ma2018sparse,wang2018plug}  using MAE, RMSE and newly proposed OMAE.
Results are summarized in Tab. \ref{tab:nuscnes}.
To this end, it is noticeable that our method outperformed other baselines on MAE 0-80 meters, \emph{e.g.}, \textbf{3713.6 (RC-PDA)} \textit{\textbf{vs}} \textbf{2467.7 (DORN)}  \textit{\textbf{vs}} \textbf{2179.3 (RadarNet)} 
 \textit{\textbf{vs}} \textbf{1927.0 (Ours)}.
Our inference speed was also the fastest among the above methods in terms of FPS, \emph{e.g.}, \textbf{ 10.4 (RC-PDA)} \textit{\textbf{vs}} \textbf{ 3.7 (RadarNet)} \textit{\textbf{vs}} \textbf{21.7 (Ours)}.   
We attribute the efficient inference speed to the advantages of accurate sparse supervision, which enables the design of our single-stage network to be smaller yet perform better.
In contrast, several previous methods \cite{long2021radar,gasperini2021r4dyn,lin2020depth,ma2018sparse,wang2018plug} utilized multiple radar frames or multiple LiDAR frames for depth completion, which inevitably introduced noises to the training phase, hindering further performance improvements on both accuracy and speed.
As for our proposed framework, we only used the single-frame radar and camera input, with the clean single-frame sparse LiDAR supervision, which achieved the SOTA accuracy and speed.

More qualitative results are shown in Fig. \ref{fig:compare}.
It is obvious that our predicted depth maps had superior visual performance compared to other methods.
For instance, in rows 1-2, the scanning pattern of RadarNet \cite{singh2023depth} was particularly severe in the night environment, making the shape of objects in the scene hard to recognize.
The depth maps predicted by DORN \cite{lo2021depth} contained grid artifacts.
In contrast, our predicted depth maps had more accurate object shape perceptions, which may benefit from the introduction of the Radar-aware Mask Decoder. 

\noindent \textbf{Ablation Studies on the \textit{Compensation} part.} Results are summarized in Tab. \ref{tab:decoder}, demonstrating that after gradually adding each module of the \textit{Compensation} part, the model achieved better and better performance. 

\begin{table}[t]
 \caption{\textbf{Effectiveness of the \textit{Compensation} part}. All experiments were conducted under sparse supervision with the proposed disruptions added. }
    \centering
    \setlength\tabcolsep{10pt}
    \renewcommand{\arraystretch}{1.0}
    \resizebox{0.9\textwidth}{!}{ 
    \begin{tabular}{c c | c c c}
    \toprule
      Radar-Position Injection Module & Radar-aware Mask Decoder 
& OMAE 0-80m (mm) & MAE 0-80m (mm) $\downarrow$ & RMSE 0-80m (mm) $\downarrow$ \\
     \midrule
      $\times$ & $\times$        & 2104.3 & 2201.1 & 5103.3 \\
      $\checkmark$ & $\times$        & 2033.9 & 2116.4 & 4970.5 \\
      $\checkmark$ & $\checkmark$    & \textbf{1817.0} & \textbf{1927.0} & \textbf{4609.6} \\
      \bottomrule
    \end{tabular}}
    \label{tab:decoder}
\end{table}

\begin{table}[t]
\caption{\textbf{Retraining LiDAR-camera depth completion methods on the radar-camera depth completion task.} Experimental results show that directly transferring the depth completion methods of LiDAR-camera input to radar-camera input caused a sharp drop in accuracy and was not suitable for our task.
    }
    \centering
    \setlength\tabcolsep{8pt}
    \renewcommand{\arraystretch}{1.0}
    \resizebox{0.75\textwidth}{!}{ 
    \begin{tabular}{cc | c c c}
    \toprule
      Original Input & Method & OMAE 0-80m (mm)  & MAE 0-80m (mm) $\downarrow$ & RMSE 0-80m (mm) $\downarrow$ \\
     \midrule
      LiDAR-camera & CompletionFormer \cite{zhang2023completionformer}  & 3088.5 & 3100.7 & 6285.8 \\
      LiDAR-camera & DySPN \cite{lin2022dynamic}      & 3452.1  
 & 3549.2 & 6801.9 \\
      \hline
      Radar-camera & Ours & \textbf{1817.0} & \textbf{1927.0} & \textbf{4609.6} \\
      \bottomrule
    \end{tabular}}
    \label{tab:lidar-camera}
\end{table}

\begin{table}[t]
 \caption{\textbf{Aligning backbones of dense supervision methods with our framework.}
 The \textit{GT} column is the same as the one in Tab. \ref{tab:nuscnes}.
  Experimental results demonstrate the negative impact of the noised dense supervision.
    }
    \centering
    \setlength\tabcolsep{8pt}
    \renewcommand{\arraystretch}{1.0}
    \resizebox{0.75\textwidth}{!}{ 
    \begin{tabular}{l c |c|c|c}
    \toprule
         & GT &  OMAE 0-80m (mm) & MAE 0-80m (mm) $\downarrow$ & RMSE 0-80m (mm) $\downarrow$ \\
     \midrule
      RadarNet \cite{singh2023depth} & 161 & 3076.7 & 3104.5 & 6291.1 \\
      DORN \cite{lo2021depth} & interpolation & 2775.0 & 2922.3 & 5895.6 \\
      RC-PDA \cite{long2021radar} & 25 & 2924.6 &  2820.1 & 6096.6  \\
      \hline
      Disruption-Compensation & 1  & \textbf{1817.0} & \textbf{1927.0} & \textbf{4609.6}\\
      \bottomrule
    \end{tabular}}
    \label{tab:dense-GT}
\end{table}

\noindent \textbf{Comparisons with LiDAR-Camera Depth Completion Methods.} To gain the full picture, we further reproduced several SOTA LiDAR-camera depth completion methods \cite{zhang2023completionformer,lin2022dynamic} to verify whether they can be directly used for the task of radar-camera depth completion. 
Specifically, we did not modify their models but only retrained on our sparse LiDAR supervision data for radar-camera depth completion.
As shown in Tab. \ref{tab:lidar-camera}, the direct translations of these methods to the radar-camera depth completion task achieved unsatisfying performance.
We argue that compared to LiDAR input data, radar input data for depth completion presents much more difficult and unique challenges, \emph{e.g.}, the height uncertainty and extreme sparsity of radar point clouds. 

\noindent \textbf{Aligning Backbones of Dense Supervision Methods.} To demonstrate the negative impact of existing dense supervision methods more fairly, which involved stacking multiple LiDAR frames or interpolation algorithms, we further conduct experiments to train their models with the same backbone of our framework, \emph{i.e.}, ResNeXt \cite{xie2017aggregated}, but still under their noised dense  supervision.
Results are shown in Tab. \ref{tab:dense-GT}, where our method under sparse supervision still outperformed previous methods, indicating the toxic effect of noised dense supervision.

\section{Conclusion}
In this paper, we have revisited the task of radar-camera depth completion and presented a new method with \textit{\textbf{sparse LiDAR supervision}} to outperform previous \textit{\textbf{dense LiDAR supervision methods}} in both accuracy and speed.
Specifically, we have revealed that the main challenge of the sparse supervision is the unexpectedly learned LiDAR distribution, termed as the \textit{LiDAR Distribution Leakage} (LDL) in this paper.
Building on this insight, we have designed a novel \textit{Disruption-Compensation} radar-camera depth completion framework to address the issue of LDL. 
Extensive experimental results have demonstrated that our proposed framework under \textit{\textbf{sparse supervision}} outperformed the SOTA \textit{\textbf{dense supervision}} methods in terms of MAE (Mean Absolute Error), OMAE (Object-level Mean Absolute Error) and FPS (Frame Per Second).
In summary, we relight the sparse supervision in the radar-camera depth completion task and demonstrate that our framework under the \textit{\textbf{accurate sparse supervision}} outperforms previous methods under \textit{\textbf{noisy dense supervision}}. 
In fact, obtaining accurate and dense supervision is a notoriously challenging task in many fields, such as depth estimation, feature matching and 3D reconstruction, prompting us to rethink the value of sparse supervision.
We hope that our proposed framework design under sparse supervision can provide more insights into these fields.


%
%
\bibliographystyle{splncs04}
\bibliography{main}
\end{document}